\documentclass[lettersize, journal]{IEEEtran}
\IEEEoverridecommandlockouts
\usepackage{makecell}
\usepackage{multirow}
\usepackage{booktabs}
\usepackage{threeparttable}

\usepackage{colortbl}
\usepackage{hhline}
\usepackage{color}
\usepackage{amsmath,amsfonts}
\usepackage[dvipsnames, svgnames, x11names]{xcolor}

\usepackage{algorithm}
\usepackage[english]{babel}
\usepackage{vcell}
\usepackage{tablefootnote}
\usepackage[utf8]{inputenc}

\usepackage{caption}

\usepackage{algorithmic}

\makeatletter
\newcommand{\removelatexerror}{\let\@latex@error\@gobble}
\makeatother

\usepackage{array}
\renewcommand\arraystretch{1.3} % double the height of table

\usepackage[caption=false,font=normalsize,labelfont=sf,textfont=sf]{subfig}
\usepackage{textcomp}
\usepackage{stfloats}
\usepackage{verbatim}
\usepackage{graphicx}
\usepackage{epsfig}
\usepackage{hyperref}
\DeclareUnicodeCharacter{2061}{}
\usepackage[backend=bibtex,style=ieee,sorting=none,isbn=false,url=false,doi=false]{biblatex}
\title{Theoretical Model Construction of Deformation-Force for Soft Grippers Part II: Displacement Control Based Intrinsic Force Sensing}

\author{Huixu Dong, Ziyi Zheng, Haotian Guo, Sihao Yang, Chen Qiu, Jiansheng Dai, I-Ming Chen, ~\IEEEmembership{Fellow,~IEEE}
\thanks{\noindent Huixu Dong, Ziyi Zheng, Haotian Guo, Sihao Yang are with Robot Perception
and Grasp Laboratory(Grasp Lab), Zhejiang University, Hangzhou 310058,
China (e-mail: huixudong@zju.edu.cn). I-Ming Chen is with Robotics Research Center, Nanyang Technological University, Singapore 639798. Chen Qiu is with Maider Medical Industry Equipment Co., Itd, China 317607. Jiansheng Dai is with Shenzhen Key Laboratory of Biomimetic Robotics and Intelligent Systems, SUSTech Institute of Robotics, Southern University of Science and Technology, Shenzhen, 518055, China, and Centre for Robotics Research, Department of Engineering, King's College London Strand, London WC2R 2LS, UK.}}
\begin{document}
\maketitle
\begin{abstract}
Force-aware grasping is an essential capability for most robots in practical applications. Especially for compliant grippers, such as Fin-Ray gripper, it still remains challenging to build a bidirectional mathematical model that mutually maps the shape deformation and contact force. Part I of this article has constructed the force-displacement relationship for design optimization through the co-rotational theory. In Part II, we further devise a displacement-force mathematical model, enabling the compliant gripper to precisely estimate contact force from deformations sensor-free. The presented displacement-force model elaborately investigates contact forces and provides force feedback for a force control system of a gripper, where deformation appears as displacements in contact points.
Afterwards, simulation experiments are conducted to evaluate the performance of the proposed model through comparisons with the finite-element analysis (FEA) in Ansys. Simulation results reveal that the proposed model accurately estimates contact force, with an average error of around 3\% and 4\% for single or multiple node cases, respectively, regardless of various design parameters (Part I of this article is released in Arxiv\footnote{Part I: \href{https://arxiv.org/pdf/2303.12987.pdf}{https://arxiv.org/pdf/2303.12987.pdf}}). 
\end{abstract}


\section{Introduction}

\begin{figure}[htbp]
\centerline{\includegraphics[width=0.85\columnwidth]{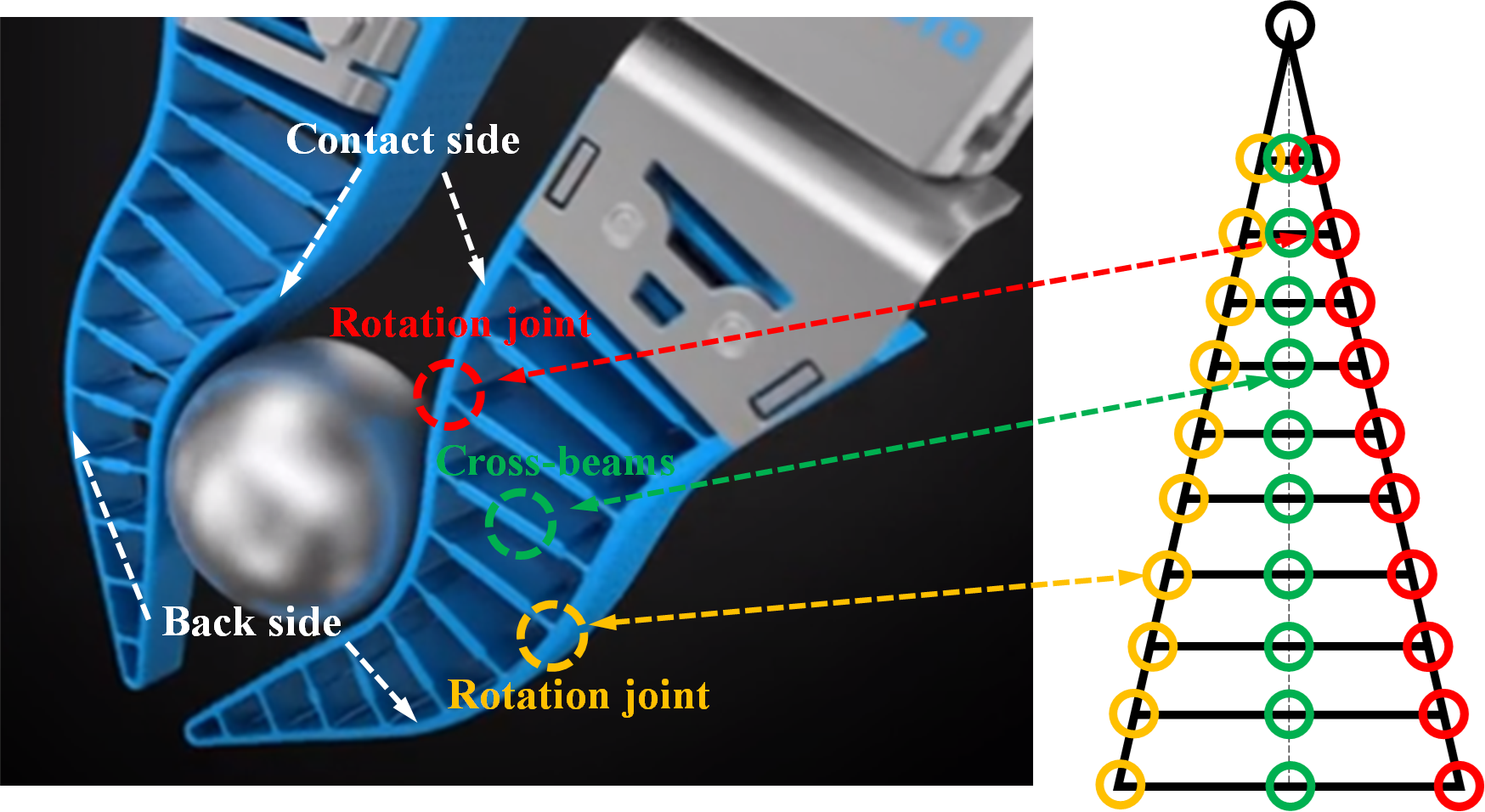}}
\vspace{-2mm}
\caption{\small{Festo’s soft gripper based on Fin-Ray® effect and its equivalent fin-ray structure, \textcolor{red}{retrived from \href{https://www.festo.com/}{Festo Gmbh}.}  The green circles represent the crossbeams; the yellow circles indicate the joints of the back side and the red circles are the joints of the contact side.}}
\vspace{-10pt}
\vspace{-2mm}
\label{fig:1}
\end{figure}

\IEEEPARstart{C}{ompliant} grasping plays an important role in robots' practical applications \cite{ref1,ref2,ref59}.  Despite significant progress in the field of compliant grasping, accurately sensing the forces involved in such grasps remains challenging as evidenced by existing literature \cite{ref2012grasping,refdong2022gsg,refdong2022construction}. Particularly, inadequate grasping forces may negatively influence grasping stability, even resulting in task failure. Meanwhile, excessive forces may cause damage to objects, especially fragile or soft ones. Therefore, it is well worth achieving the interaction forces between grippers and objects. As introduced in Part I, in this article we choose the Fin-Ray gripper as an example (see Fig. \ref{fig:1}) for its generality to other compliant grippers.

For a compliant grasp, it poses a considerable challenge to obtaining the contact forces. Some force sensors that rely heavily on force-sensitive resistors, strain gauges, and electrodes were developed and integrated into compliant grippers \cite{ref34,ref35}. However, they are not suitable for when large strain at finger is a need. A few new sensors appeared for obtaining contact forces of soft grippers based on the characteristics of capacitance or optical structures \cite{ref36,ref37}, but the complexity of compliant grippers increases accordingly.
Besides, there exist  good solutions for achieving contact forces through theoretical models of deformations and contact forces \cite{ref4,ref6,ref38,ref39}. Such a model revealing the deformation-force mathematical relationships is established by virtual work \cite{ref38}, or piecewise constant curvature and Cosserat-rod theory \cite{ref39,ref46}, or Newton–Euler iterations \cite{ref47}, or two recent methods based on machine learning \cite{ref6} and improved pseudo-rigid body model with the virtual work \cite{ref4}. The virtual work method \cite{ref38, ref43}, piecewise constant curvature and Cosserat-rod theory \cite{ref39, ref46, ref50}, Newton–Euler iterations \cite{ref47} typically focus on continuum robots with long thin structures while applications of such modelling methods on compliant grippers have not been reported, that is, outside of the field of continuum robotics, these methods are uncommon. Meanwhile, these robotic bodies are considered as a curve, whose cross-sections have the same rotations; however, the contact side and back of a compliant finger present significantly different deformations with non-uniformity joints.

\textcolor{Green}{Xu et al. \cite{ref6} introduced a neural network method to achieve the mapping relationships from the deformations to contact forces, skillfully avoiding building the corresponding model of compliant grasp. However, the supervised learning method yields two drawbacks, resulting in limited applications in real scenarios. In particular, the performance of the learning-based model highly relevant to the collected data, both quality and quantity. With the change in the geometrical parameters of the gripper, the model will be ineffective and new data needs to be collected, with repeated effort required.} In the meantime, they insert many rigid rods to modify the gripper structure, which improves gripper structure stiffness for the static model to be linear. By contrast, this also indicates that method is only suitable for linear-structured fin-ray grippers. Shan et al. \cite{ref4} proposed a theoretical model for deformation-force via an improved pseudo-rigid body model with the virtual work. However, this method's stiffness matrix assumes identical transformation frames or minor changes, the model cannot handle a scene where compliant fingers cause significant deformation. Shan's method can just tackle symmetrical-structure-compliant grippers, which is limited to broad applications. Moreover, these two methods are constrained to compliant grippers with fin-ray effects and rely on numerous assumptions for the deformation-force relationship.

To fill the above research gap, we seek to further construct the deformation-force model, providing crucial insights into design optimization and force control of fin-ray grippers. First, the proposed methods, based on the co-rotational theory, can map contact forces from deformations. The displacement-force modeling estimates contact forces, which potentially facilitates the force control of a gripper. Second, a series of comparison simulations based on the proposed model and the standard FEA in Ansys is implemented to systematically evaluate the performance of the proposed model.
% In particular, a family of fin-ray grippers with various design parameters are designed to investigate the optimal design of a fin-ray gripper. Then, when single-node or multiple-node contact occurs, simulation experiments are carried out to evaluate the performance of the proposed model in estimating contact forces. \textcolor{cyan}{Finally, after determining the design parameters optimized by the proposed method, we build a fin-ray gripper. Furthermore, we experimentally demonstrate that the proposed model can accurately predict both the contact forces and the overall grasp force.}

% We highlight the novelties of our work. Our core contribution is constructing the deformation-force mathematical models, giving the first generalizable deformation-force solution for compliant rigid and soft grippers with various structures and materials under limited assumptions, even when grippers deform significantly. The second novelty involves modeling and analyzing the mathematical relationships between finger deformation and contact forces using co-rotational theory. We are the first to establish the deformation-to-force mapping model, providing a critical insight into grasping force control by sensing contact forces in simulations. 

We highlight the novelties of this work. \textbf{Firstly}, this article, for the first time, explores the theoretical construction of the deformation-force relationship based on physical models through co-rational concepts. \textbf{Secondly}, the individual influence of the co-rotational modeling parameters, namely mesh density and node radius factor, on the model's accuracy is systematically investigated. 
% the proposed mathematical model can be easily generalized to other compliant grippers, such as compliant rigid and soft grippers with few assumptions made. 
% \textbf{Thirdly}, the comparison with FEA in simulation is innovatively adopted as \textcolor{Purple}{FEA has poor real-time force control and cannot apply to the displacement control of our gripper.} 
This article will not only endow compliant grippers with a powerful mathematical tool that provides force perception, laying the groundwork for further research into theoretical model construction, but inspire more systematical and optimized gripper design.

\vspace{-1mm}
\section{modeling and analysis}

\subsection{Co-rotational Model}

As demonstrated in Part I of this article, we introduce the co-rotational modeling following Crisfield \cite{ref51} and Yaw \cite{ref52}, accurately and efficiently modeling any arbitrary large motion of objects. Then, two algorithms are proposed, where \textbf{Algorithm 1} acquires the current pose of beam element (current length \textbf{$L$}, the incline angle $\beta$, together with local internal force vector $\textbf{\textit{q}}_l$ and internal global force vector $\textbf{\textit{F}}_{int}$), and \textbf{Algorithm 2} updates the modified global tangent stiffness matrix $\textbf{\textit{K}}_s$ respectively. Besides, the force-deformation model, \textbf{Algorithm 3}, is devised for systematical design optimization. Considering nodes or other hinge-type connections between the separate beam elements \cite{ref53} in a real-world application, we introduce an index of effective length modification $R_m$ and thus,
\vspace{-2.5mm}
\begin{equation}
\begin{matrix}
L_0=L_0-2R_m\\
L=L-2R_m  
\end{matrix}
\label{eq:6}
\vspace{-2mm}
\end{equation}
\noindent The value of $R_m=\mu R_{node}$ is determined in terms of the connection shapes, with $R_{node}$ the connection node radius and $\mu$ the modification factor, determined by comparison with the standard finite element simulation.

\vspace{-3mm}
\subsection{Displacement-Force Modeling}

% The force control algorithm given in the last section is satisfactory for many problems, for instance, when the external force is given to calculate the deformation of the gripper. However, in the case when a target deformation is given to estimate the reaction force, the force control algorithm may not well solve the problem. In addition, when problems of snap-through, snap-back, or equilibrium path tracing are encountered, force control algorithm will not be able to fully describe the system’s force-displacement behaviour. To solve this problem, displacement control algorithms [51, 55] are needed. 
In this section, a displacement-force relationship closely following controls schemes from McGuire et al. \cite{ref57} and Clarke and Hancock \cite{ref58} is provided, where a constantly increased external displacement load is applied at the target structure instead of an external force. 
% In addition, because of the nature of constantly increased displacement, the “snap-through” problem can be solved. 

The details of the proposed modeling are as follows. Each node has three DoFs, including x-axis and y-axis translations and a rotation. If the maximum displacement for the x-axis or y-axis translation or the rotation is defined as a vector $\textbf{\textit{D}}_{total}$, and a total number $n_{inc}$ of increments are required to reach the final equilibrium, we have:
\vspace{-2mm}
\begin{equation}
\Delta \bar{\textbf{\textit{u}}}_f = \frac{\textbf{\textit{D}}_{total}}{n_{inc}} 
    \label{eq:29}
    \vspace{-2mm}
\end{equation}

\noindent where $\Delta \bar{\textbf{\textit{u}}}_f$ represents an incremental value.

 A full equilibrium cycle in the $n$-th increment is demonstrated. We obtain the modified stiffness matrix $\textbf{\textit{K}}_s$ according to \textbf{Algorithm 1} and \textbf{Algorithm 2} based on the initial values resulting from the $(n-1)$-th cycle. Then, a global nodal displacements vector can be calculated as:
% As demonstrated in our prior work [arxiv], where any arbitrary large motion of objects can be accurately and efficiently modeled following the concept of the co-rotational modeling given by Crisfield[51] and Yaw[52], two  algorithms are proposed, where algorithm 1 acquires the current pose of beam element, together with global/local force vectors, and algorithm 2 updates the modified global tangent stiffness matrix $\textbf{\textit{K}}_s$ respectively
% We obtain the modified stiffness matrix $\textbf{\textit{K}}_s$ via \textbf{Algorithm \ref{alg:1}} and \textbf{Algorithm \ref{alg:2}} based on the $(n-1)$-th cycle's initial values. 
% Then, we calculate a global nodal displacements vector as
\vspace{-2mm}
\begin{equation}
\hat{\textbf{\textit{u}}}=\textbf{\textit{K}}_s^{-1}\textbf{\textit{F}}_{ref}
\label{eq:30}
\vspace{-2mm}
\end{equation}

\noindent where $\textbf{\textit{F}}_{ref}$ is the referenced value \cite{ref56}.
% rather than the total value of global nodal forces $\textbf{\textit{F}}_{total}$ used in the force control. 
The value of $\textbf{\textit{F}}_{ref}$ is determined according to either experience or an automatic strategy. In our case, it is found that the direction of $\textbf{\textit{F}}_{ref}$ is more important than its magnitude. As a result, the direction of $\textbf{\textit{F}}_{ref}$ is set the same as: $\textbf{\textit{D}}_{total}$. For the magnitude, we use a try-and-error approach. 
% by submitting a pre-defined $\textbf{\textit{F}}_{ref}$ into the force-control algorithm so that the resulted displacement is with the same order of magnitude of $\textbf{\textit{D}}_{total}$. 
Then we can obtain the incremental load ratio vector $d\boldsymbol{\lambda}^{n+1}$ as:
\vspace{-1mm}
\begin{equation}
d\boldsymbol{\lambda}^{n+1} = \frac{\Delta \bar{\textbf{\textit{u}}}_f}{\hat{\textbf{\textit{u}}}} 
\label{eq:31}
\vspace{-1mm}
\end{equation}
\noindent And then update the load ratio \cite{ref58}: 
\vspace{-1mm}
\begin{equation}
\boldsymbol{\lambda}^{n+1}=d\boldsymbol{\lambda}^{n+1}+\boldsymbol{\lambda}^{n}  
\label{eq:32}
\vspace{-2mm}
\end{equation}
\noindent Thus, the incremental force vector can be given as:
\vspace{-1mm}
\begin{equation}
d\textbf{\textit{F}} = d\boldsymbol{\lambda}^{n+1}\textbf{\textit{F}}_{ref}
\label{eq:33}
\vspace{-2mm}
\end{equation}
\noindent We calculate the incremental global nodal displacements as:
\vspace{-1mm}
\begin{equation}
d\textbf{\textit{u}} = \textbf{\textit{K}}^{-1}_sd\textbf{\textit{F}}
\label{eq:34}
\vspace{-2mm}
\end{equation}                           
\noindent Then the global nodal displacements and global nodal forces can be updated as:
\vspace{-4mm}
\begin{equation}
\vspace{-1mm}
\begin{matrix}
    \textbf{\textit{u}}^{n+1} = \textbf{\textit{u}}^n +d\textbf{\textit{u}} \\
    \textbf{\textit{F}}^{n+1} = \textbf{\textit{F}}^n +d\textbf{\textit{F}}
\end{matrix}
\label{eq:35}
\vspace{-2mm}
\end{equation}

\noindent Further, we can update $\textbf{L}, \textbf{c}, \textbf{s}, \textbf{q}_l^{n+1}, \!\textbf{F}_{int}^{n+1}$ according to \textbf{Algorithm 1} based on $\textbf{u}^{n+1}$. By accounting for the required support, we calculate the residual as 
\vspace{-1mm}
\begin{equation}
    \textbf{\textit{R}} = \boldsymbol{\lambda}^{n+1}\textbf{\textit{F}}_{ref} - \textbf{\textit{F}}^{n+1}_{int}
\label{eq:36}
\vspace{-2mm}
\end{equation}

\noindent The norm of the residual is provided as
\begin{equation}
    R = \sqrt{\textbf{\textit{R}}\cdot \textbf{\textit{R}}}
    \label{eq:37}
    \vspace{-2mm}
\end{equation}

\noindent Subsequently, we implement an iterative strategy to obtain the final values of the node displacement and externally applied force. A few iteration variables are defined as follows: Iteration number $k=0$, $tolerance=10^{-3}$. The maximum iteration step is limited by $maxiter=100$. The correction to incremental global nodal displacements is defined as $\delta \textbf{\textit{u}}^k (\delta \textbf{\textit{u}}^0=\textbf{0})$, and the correction to load ratio is defined as $\delta \boldsymbol{\lambda}^k (\delta \boldsymbol{\lambda}^0=\textbf{0})$. The storage vector of local forces in this iteration cycle is set as $\textbf{\textit{q}}_{l-temp}^k=\textbf{\textit{q}}_l^{n+1}$.

\begin{figure*}[t]
\vspace{-1mm}
\centerline{\includegraphics[width=1.8\columnwidth]{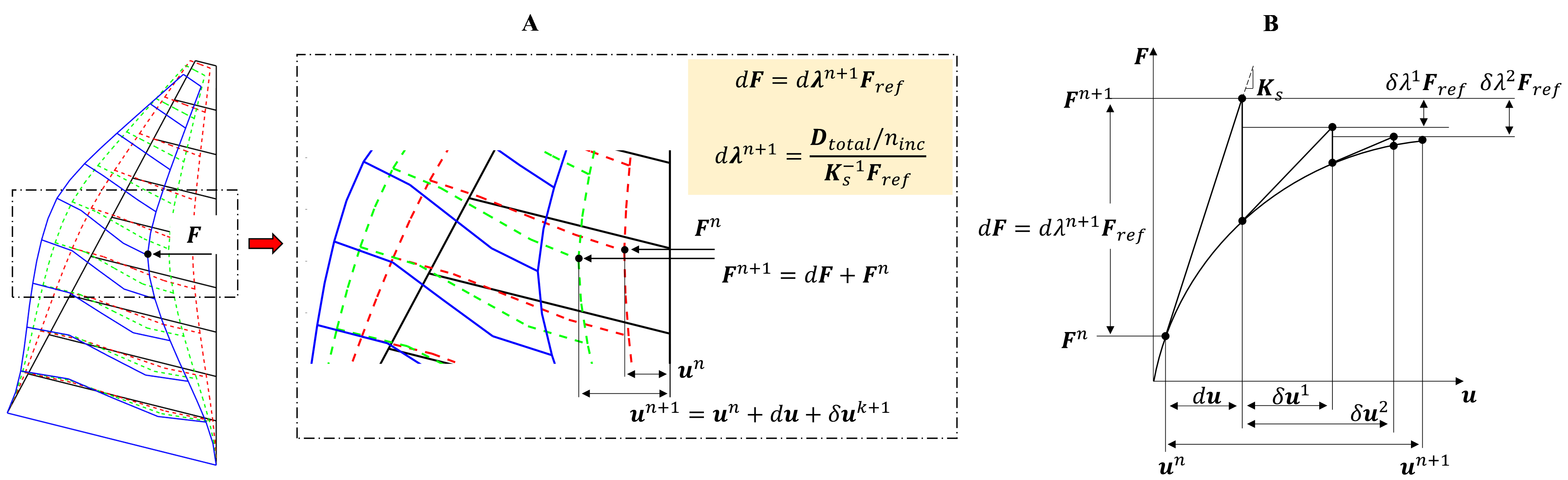}}
\vspace{-2mm}
\caption{\small One incremental step in the displacement control. The sketch of finger deformation(A), the relationship of force and deformation(B).}
\label{Fig:6}
\vspace{-4mm}
\end{figure*}

In each step $k$, the stiffness matrix $\textbf{\textit{K}}_s$ is updated according to \textbf{Algorithm 2} based on updated current values of inputs and $\textbf{\textit{q}}_{l-temp}^k$. The load ratio is calculated following \cite{ref56} as
\vspace{-2mm}
\begin{equation}
    \begin{matrix}
        \acute{\textbf{\textit{u}}} = \textbf{\textit{K}}_s^{-1}\textbf{\textit{R}} \\
        \grave{\textbf{\textit{u}}} = \textbf{\textit{K}}_s^{-1}\textbf{\textit{F}}_{ref}\\
        \delta \boldsymbol{\lambda }^{k+1} = \delta\boldsymbol{\lambda }^k - \frac{\acute{\textbf{\textit{u}}}}{\grave{\textbf{\textit{u}}}} 
    \end{matrix}
    \label{eq:38}
    \vspace{-2mm}
\end{equation}
Then the correction to $\textbf{\textit{u}}^{n+1}$ can be calculated as
\vspace{-1mm}
\begin{equation}
\delta \textbf{\textit{u}}^{k+1} = \delta \textbf{\textit{u}}^{k} + \textbf{\textit{K}}_s^{-1}
\begin{bmatrix}
    \textbf{\textit{R}}-\frac{\acute{\textbf{\textit{u}}}}{\grave{\textbf{\textit{u}}}}\textbf{\textit{F}}_{ref}
\end{bmatrix}
    \label{eq:39}
    \vspace{-2mm}
\end{equation}
Update $\textbf{\textit{q}}_{l-temp}^{k+1}$ and $\textbf{\textit{F}}_{int}^{n+1}$ according to \textbf{Algorithm 1}, and the residual in this iteration can be calculated as
\vspace{-1mm}
\begin{equation}
    \textbf{\textit{R}} = (\boldsymbol{\lambda}^{n+1}+\delta\boldsymbol{\lambda^{k+1}})\textbf{\textit{F}}_{ref}-\textbf{\textit{F}}^{n+1}_{int}
    \label{eq:40}
    \vspace{-2mm}
\end{equation}
The norm of the residual is updated using  Eqn.(\ref{eq:37}).
% \vspace{-1mm}
% \begin{equation}
%     R = \sqrt{\textbf{\textit{R}}\cdot \textbf{\textit{R}}}
%     \label{eq:41}
%     \vspace{-2mm}
% \end{equation}

Then updating iteration number $k=k+1$, the iteration cycle will terminate if $R\le tolerance$ or $k\ge maxiter $ (in this case, the convergence criteria are not met), and the variables will update to their final value in this $n$-th increment
\vspace{-2mm}
\begin{equation}
\begin{matrix}
    \boldsymbol{\lambda}^{n+1} = \boldsymbol{\lambda}^{n+1} + \delta\boldsymbol{\lambda}^{k+1} \\
    \textbf{\textit{q}}_{l}^{n+1} = \textbf{\textit{q}}_{l-temp}^{k+1} \\
    \textbf{\textit{u}}^{n+1} = \textbf{\textit{u}}^{n} + d\textbf{\textit{u}}+\delta \textbf{\textit{u}}^{k+1}
\end{matrix}
    \label{eq:42}
    \vspace{-2mm}
\end{equation}

\begin{algorithm}[h]
\renewcommand{\thealgorithm}{4}
\caption{\textit{\textbf{Displacement-force Modeling}}}
\begin{algorithmic}
\STATE \textbf{\textit{Input:}}
\STATE \small$\!\!n_{nodes},n_{mem}, m_{conn},\textit{\textbf{A,E,I}}, \textit{\textbf{x}}_0, \textit{\textbf{y}}_0, \textit{\textbf{L}}_0, \!\boldsymbol{\beta}_0, \textit{\textbf{u}},R_m,\textbf{\textit{D}}_{total}, \textbf{\textit{F}}_{ref}$
\STATE \textbf{\textit{Calculations:}}
\STATE \textbf{\textit{For}} $n$=1: $n_{mem}$

\STATE \quad\textit{calculate $\Delta \bar{\textbf{u}}_f$ by Eqn.(\ref{eq:29})}

\STATE \quad\textit{calculate $\hat{\textbf{u}}$ by Eqn.(\ref{eq:30})}

\STATE \quad\textit{calculate $d\boldsymbol{\lambda}^{n+1},\boldsymbol{\lambda}^{n+1}$ by Eqns.(\ref{eq:31}-\ref{eq:32})}

\STATE \quad\textit{calculate $d\textbf{F}$ by Eqn.(\ref{eq:33})}

\STATE \quad\textit{calculate $\textbf{L}, \textbf{c}, \textbf{s}, \textbf{q}_l, \textbf{F}_{int}$ by \small{\textbf{Algorithm 1}}}

\STATE \quad\textit{calculate $\textbf{K}_s$ by \textbf{Algorithm 2}}

\STATE \quad\textit{calculate $d\textbf{u}$ by Eqn.(\ref{eq:34})}

\STATE \quad\textit{update $\textbf{u}^{n+1}$ and $\textbf{F}^{n+1}$ by Eqn.(\ref{eq:35})}

\STATE \quad\textit{update $\textbf{L}, \textbf{c}, \textbf{s}, \textbf{q}_l^{n+1}, \!\textbf{F}_{int}^{n+1}$ \!by \small{\textbf{Algorithm 1}} \normalsize{based on $\textbf{u}^{n+1}$}}

\STATE \quad\textit{calculate the residual $\textbf{R}$ and $R$ by Eqns.(\ref{eq:36}-\ref{eq:37})}

\STATE \quad\textit{set up iteration variables $k, tolerance, maxiter, \delta \textbf{u}, \delta \boldsymbol{\lambda}$ and $ \textbf{q}_{l-temp}^k$  }

\STATE \quad\textit{
start iterations while $R\! \ge\! tolerance$ and $k\! \le \!maxiter$}

\STATE \qquad\enspace\textit{\footnotesize{\expandafter{\romannumeral1}. calculate $\textbf{K}_s$ by \textbf{Algorithm 2}}}

\STATE \qquad\enspace\textit{\footnotesize{\expandafter{\romannumeral2}. update load ratio correction $\delta \boldsymbol{\lambda}^{k+1}$  by Eqn.(\ref{eq:38})}}

\STATE \qquad\enspace\textit{\footnotesize{\expandafter{\romannumeral3}. update $\delta\textbf{u}^{k+1}$ by Eqn. (\ref{eq:39})}}

\STATE \qquad\enspace\textit{\footnotesize{\expandafter{\romannumeral4}. update $\textbf{q}_{l-temp}^{k+1},\textbf{F}_{int}^{n+1}$ by \textbf{Algorithm 1}}}

\STATE \qquad\enspace\textit{\footnotesize{\expandafter{\romannumeral5}. 
calculate the residual $\textbf{R}$ by Eqn.(\ref{eq:40}), and its norm $R$ by Eqn.(\ref{eq:37})}}

\STATE \qquad\enspace\textit{\footnotesize{\expandafter{\romannumeral6}. update iteration number k=k+1
}}

\STATE \quad\textit{End of while loop iterations}

\STATE \quad\textit{Update variables $\boldsymbol{\lambda}^{n+1}, \textbf{q}_l^{n+1}$ and $\textbf{u}^{n+1}$ by Eqn. (\ref{eq:42})}

\STATE \textbf{\textit{End}}     
\STATE \textit{\textbf{Output:} $\boldsymbol{\lambda}^{n+1},\textbf{q}_l^{n+1},\textbf{u}^{n+1}$}
\end{algorithmic}
\label{alg:4}
\end{algorithm}
\vspace{-1mm}

The complete displacement control in the $n$-th increment is also illustrated in Fig. \ref{Fig:6}, where the variables updated in both the preliminary step and iteration cycle are demonstrated. The detailed algorithm is shown in \textbf{Algorithm 4}.

\vspace{-2mm}
\section{simulation experiments}

Simulation experiments were conducted to evaluate the performance of the proposed co-rotational approach by comparing it with the finite-element analysis (FEA), which is an important benchmark solution for the numerical analysis of mechanical models \cite{Force_FR}. \textcolor{Purple}{Especially for Fin-Ray grippers, research have proven its high accuracy compared with physical experiments, with an average error of around 3\%  \cite{ref4, ref6}}. Here a given compliant finger is meshed by rigid nodes, corresponding to the physical gripper in real scenarios, and detailed parameters are mentioned in Table I.
\begin{table}[thbp]
\vspace{-2mm}
\caption*{\small{Table \uppercase\expandafter{\romannumeral1}. The parameters of the sparsely and densely meshed models.}}
\vspace{-4mm}
\label{Table:1}
\begin{center}
\begin{threeparttable}[b]
\setlength{\tabcolsep}{1.8mm}{
\begin{tabular}{ccc}
\hline
Items & Sparse & Dense \\ \hline
Node number & $30$ & $66$ \\
Member & $38$ & $74$ \\
Width $m$(m) & $40e^{-3}$ & $40e^{-3}$ \\
Height $n$(m) & $80e^{-3}$ & $80e^{-3}$ \\
Node radius $R_{node}$(m) & $0.75e^{-3}$ & $0.75e^{-3}$ \\
Node radius modification factor & $1$ & $0.5$ \\
Cross section of each member $(b,h)$(m) & $20e^{-3},1e^{-3}$ & $20e^{-3},1e^{-3}$ \\
Young's modulus E (Pa) & $2e^7$ & $2e^7$ \\ \hline
\end{tabular}

}
% \begin{tablenotes}
%      \item \footnotesize{Dis$.^1$ denotes the displacement; Ave$.^2$ indicate the average value; N.A$.^3$ represents that the FE method is invalid.}
   % \end{tablenotes}
\end{threeparttable}
\end{center}
\vspace{-7mm}
\end{table}
 We  apply a pre-defined displacement load at the specified nodes (8 and 9) and estimate contact forces utilizing the proposed co-rotational theory. Similaly, the displacement load is exerted on the same finger using FEA. Fig. \ref{fig:6} reveals a good agreement between them. 
% Following the mathematical modeling, we conduct simulation experiments to evaluate the performance of proposed co-rotational approach via the comparison between the proposed model and the finite-element analysis (FEA), as=0 shown in Fig.7. FEA is considered as an important benchmark solution for the numerical analysis of mechanical models. Here a given compliant finger is meshed by nodes. Indeed, these nodes modeled as rigid elements are in accordance with the physical gripper in real scenarios. 
% % We indirectly conduct the comparison between the proposed method and FEA simulation in estimating contact force since FEA simulation cannot read contact force. 
% Specifically, contact forces can be first achieved by the proposed co-rotational approach; then, the calculated forces are exerted on the same soft finger, which allows the finger to generate displacement; finally, displacements from two methods are compared.  In terms of displacement estimation, we directly compare our method with FEA simulation since FEA can record displacements when a finger undertakes external forces. 
The rest of the section is organized as follows. 
\begin{enumerate}
\item The effects of two representative co-rotational modeling parameters, such as the mesh density parameter and node radius factor, on the model’s accuracy are investigated. 
\item We further evaluate the performance of the co-rotational approach in estimating contact forces via the proposed displacement control. 
% \item The optimization design of a soft gripper is conducted in terms of a series of key design parameters, such as the top angle of a gripper, the number and inclination angle of crossbeams, as well as the connection type between crossbeams and front \& back fin-rays.
\end{enumerate}

\begin{table*}[thbp]
\vspace{-0mm}
\renewcommand\arraystretch{0.9}
\caption*{\small{Table \uppercase\expandafter{\romannumeral2}. Results of the evaluations of single-node force estimation. The values with blue background represent the average values with different nodes at the same displacement; the values with orange background denote the average values with the same nodes at different displacements; the green-background value is the total average value; the row with grey background has the invalid data (N.A.), which is not considered when an average value is calculated. Note that all the values are considered absolute values when average values are calculated.}}
\vspace{-1mm}
\label{Table:5}
\vspace{-2mm}
\begin{center}
\begin{threeparttable}[b]
\setlength{\tabcolsep}{1.4mm}{
\begin{tabular}{c|cccccccc|cccccccc}
\hline
 & Dis$.^1$ & 2mm & 4mm & 6mm & 8mm & 10mm &  &  & Dis$.^1$ & 2mm & 4mm & 6mm & 8mm & 10mm &  &  \\ \hline
 & Node & \multicolumn{5}{c}{Error ratio(\%)} & A$.^2$(\%) & \multicolumn{1}{l|}{SD$.^3$(\%)} & Node & \multicolumn{5}{c}{Error ratio(\%)} & A$.^2$(\%) & \multicolumn{1}{l}{SD$.^3$(\%)} \\
 & \cellcolor[HTML]{9B9B9B}$1^{st}$ & \cellcolor[HTML]{9B9B9B}-4 & \cellcolor[HTML]{9B9B9B}-13 & \cellcolor[HTML]{9B9B9B}-13 & \cellcolor[HTML]{9B9B9B}-10 & \cellcolor[HTML]{9B9B9B}N.A$.^3$ & \cellcolor[HTML]{9B9B9B}10 & \cellcolor[HTML]{9B9B9B}3.67 & $6^{th}$ & -4 & -4 & -4 & -5 & -5 & \cellcolor[HTML]{F9D7AE}4.4 & \cellcolor[HTML]{F9D7AE}0.49 \\
 & $2^{nd}$ & -2 & -3 & -4 & -5 & -6 & \cellcolor[HTML]{F9D7AE}4 & \cellcolor[HTML]{F9D7AE}1.41 & $7^{th}$ & -3 & -3 & -3 & -4 & -4 & \cellcolor[HTML]{F9D7AE}3.4 & \cellcolor[HTML]{F9D7AE}0.49 \\
 & $3^{rd}$ & -4 & -4 & -4 & -5 & -5 & \cellcolor[HTML]{F9D7AE}4.4 & \cellcolor[HTML]{F9D7AE}0.49 & $8^{th}$ & 0 & 0 & 0 & 0 & 0 & \cellcolor[HTML]{F9D7AE}0 & \cellcolor[HTML]{F9D7AE}0 \\
 & $4^{th}$ & -4 & -4 & -5 & -5 & -6 & \cellcolor[HTML]{F9D7AE}4.8 & \cellcolor[HTML]{F9D7AE}0.75 & \cellcolor[HTML]{9B9B9B}$9^{th}$ & \cellcolor[HTML]{9B9B9B}8 & \cellcolor[HTML]{9B9B9B}8 & \cellcolor[HTML]{9B9B9B}9 & \cellcolor[HTML]{9B9B9B}11 & \cellcolor[HTML]{9B9B9B}N.A$.^3$ & \cellcolor[HTML]{9B9B9B}9 & \cellcolor[HTML]{9B9B9B}1.22 \\
\multirow{-6}{*}{Single-node} & $5^{th}$ & -4 & -4 & -5 & -5 & -5 & \cellcolor[HTML]{F9D7AE}4.6 & \cellcolor[HTML]{F9D7AE}0.49 & A$.^2$(\%) & \cellcolor[HTML]{ECF4FF}3.00 & \cellcolor[HTML]{ECF4FF}3.14 & \cellcolor[HTML]{ECF4FF}3.57 & \cellcolor[HTML]{ECF4FF}4.14 & \cellcolor[HTML]{ECF4FF}4.43 & \cellcolor[HTML]{D9F5B1}3.66 & \cellcolor[HTML]{D9F5B1}\textbackslash{} \\ \hline
\end{tabular}}
\begin{tablenotes}
     \item \footnotesize{Dis$.^1$ denotes the displacement; A$.^2$: the average value; N.A.$^3$ indicates that the FE method in Ansys fails; SD$.^3$: the standard deviation.}
   \end{tablenotes}
\end{threeparttable}
\end{center}
\vspace{-6mm}
\end{table*}

\vspace{-1mm}
\subsection{Co-rotational Modelling Parameter Comparison}
\vspace{-1mm}
\subsubsection{Comparison in terms of the mesh density}

\ 

\begin{figure}[H]
% \vspace{-2mm}
	\centering
 \vspace{-3mm}
	\begin{minipage}[t]{\linewidth}
		\centering
		\includegraphics[width=0.75\columnwidth]{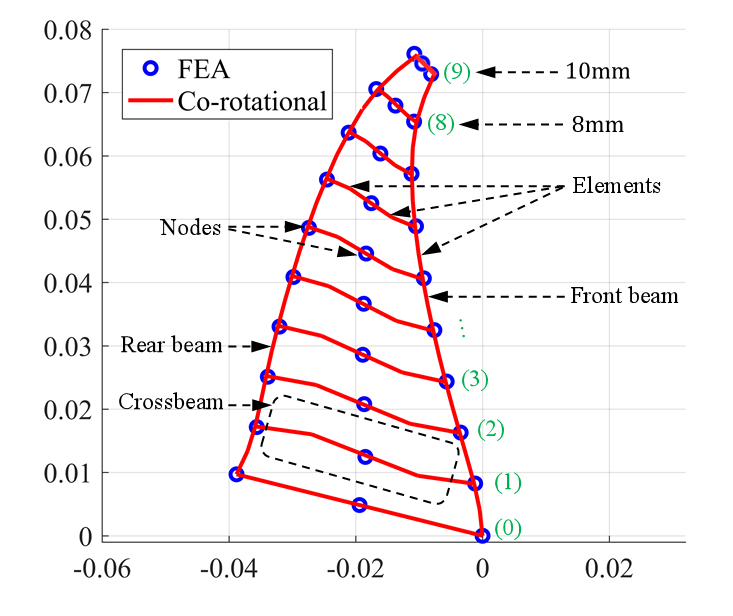}
  \vspace{-5mm}
		\caption{\small{A comparison between the co-rotational modeling and FEA simulation. In this case, external displacement loads are applied at nodes 8 and 9. The deformation shape of the soft finger is described in the red line, and the FEA simulation result of key nodes is described in the blue circle, which shows a good agreement.}}
		\label{fig:6}
  \vspace{0mm}
	\end{minipage}
	\\
	\begin{minipage}[t]{\linewidth}
		\centering
		\includegraphics[width=0.75\columnwidth]{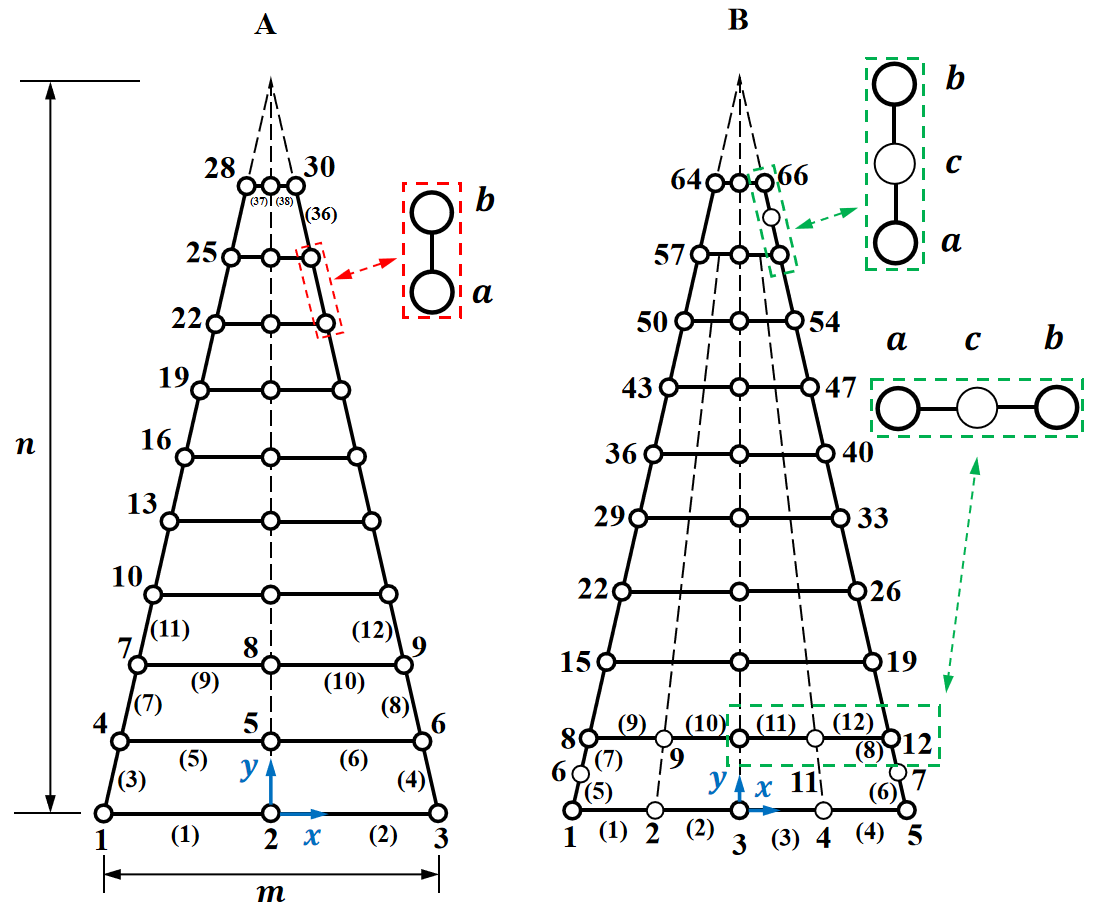}
  \vspace{-3mm}
		\caption{\small{The models meshed sparsely (A) and densely (B). m and n represent the width and height of a fin-ray finger, respectively. The numbers indicate the labels of nodes, and the numbers in brackets denote the labels of flexible elements. The red frame with two nodes, a and b, represents the sparse connection style (A), where the bold circle indicates a physically existing node. For the dense connection style, an intermediate node c is introduced, which is a virtual node used in the analytical model.}}
		\label{fig:7}
  \vspace{-2.5mm}
	\end{minipage}

\end{figure}

We first examine the effect of the number of meshed nodes on the performance of the proposed co-rotational approach. In the sparse model, we discretize the front/rear beam of a finger into nine flexible elements using ten nodes and the crossbeam into two flexible elements with one node (seen in Fig. \ref{fig:7}), generating a total of 30 nodes. By contrast, the dense model adds one additional node in the flexible element of the sparse model, resulting in altogether 68 nodes. 

\begin{figure*}[thbp]
\centerline{\includegraphics[width=1.82\columnwidth]{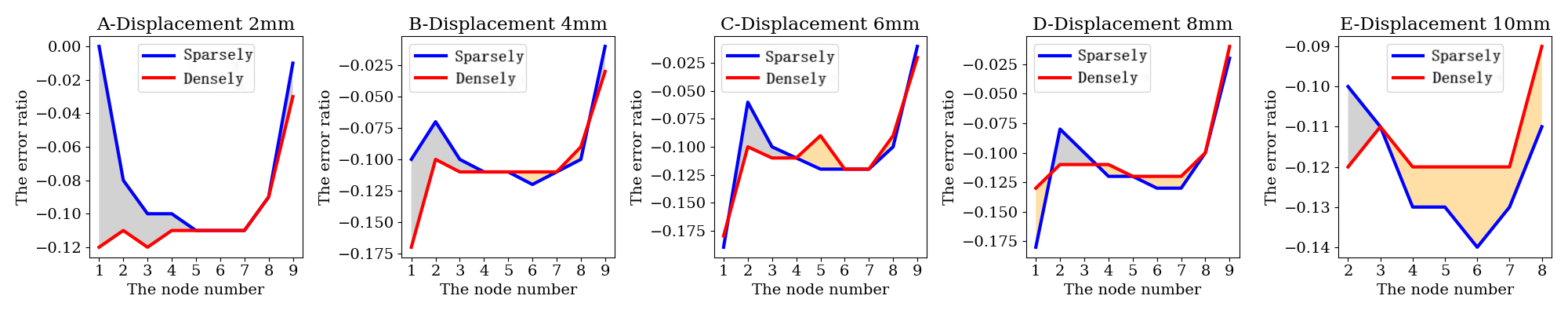}}
\vspace{-3mm}
\caption{\small{The models meshed sparsely (blue) and densely (red) in terms of mesh density.}}
\label{Fig:9}
\vspace{-2mm}
\end{figure*}

The accuracy of both sparse and dense models in estimating forces is examined. For each model, nine nodes that physically exist on the front beam of the soft finger are selected, and a displacement load ranging from 2mm to 10mm is applied. Fig. \ref{Fig:9} compares and illustrates the error ratios of both models. Generally, two models own similar error ratios at intermediate nodes (3,4,5,6,7) at various displacement loads except for the 10mm case. While, the proposed method provides poorer performance at bordering nodes (1,2 and 8,9) regardless of the load magnitudes. However, since the most commonly used nodes are 3,4,5,6,7 in practical grasping scenarios, the influence of these negative results at node 1,2,8,9 can be minimized. It is worth noting that the dense model shows more consistent error ratios compared to the sparse model regarding the varying displacement-load magnitudes at various nodes. Therefore, the co-rotational dense-meshed model is employed in the following simulations for its better performance. 

\vspace{0mm}
\subsubsection{Comparison of the node radius factor}

\ 

\begin{figure*}[thbp]
\vspace{-2mm}
\centerline{\includegraphics[width=1.82\columnwidth]{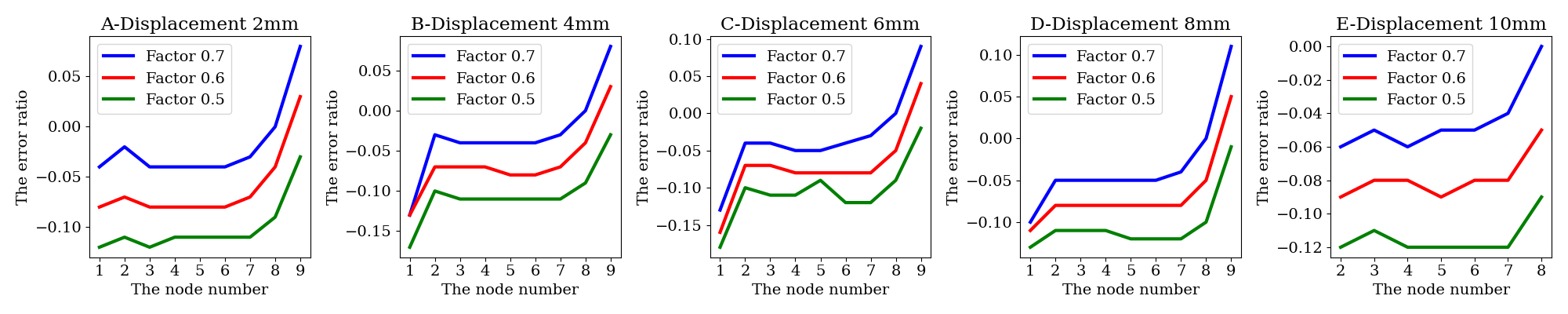}}
\vspace{-3mm}
\caption{\small{The error ratio of the densely meshed model with three node radius factors.}}
% \caption{\small{The models meshed densely for the node radius factors.}}
\label{Fig:10}
\vspace{-1mm}
\end{figure*}

The node radius affects the accuracy of the co-rotational model in estimating contact forces. Inheriting the parameters set in the above dense-meshed model, we set three factors, 0.7, 0.6, and 0.5, to adjust the node radius in the simulation. The node radius affects the effective length of each beam. In particular, larger radium indicates a smaller effective length, resulting in a larger structural stiffness. Since the displacements of FEA simulation and co-rotational approach are compared at each node, the mathematical model with a larger node radius will have a smaller displacement at each node, thus the error line is higher (error line is the combination of error ratios at nine nodes). For the displacements from 2 to 10mm with the 2mm interval, the proposed co-rotational model with a radius factor of 0.7 has better performance than others, as shown in Fig. \ref{Fig:10}. 


% \begin{figure}[htbp]
% \vspace{-2mm}
% \centerline{\includegraphics[width=0.95\columnwidth]{Fig/The deformations of a gripper for single-node displacement.png}}
% \caption{\small{The deformations of a gripper for single-node displacement loads. Displacements at nodes 1(A), 2(B), 3(C), 4(D), 5(E), 6(F), 7(G), 8(H), 9(I).}}
% \label{Fig:11}
% \vspace{-8mm}
% \end{figure}

\begin{table*}[t]
\vspace{-2mm}
\renewcommand\arraystretch{0.8}
\caption*{\small{Table \uppercase\expandafter{\romannumeral3}. Results of the evaluations of two-node force estimation. The color notations are the same as Table II. The positive error ratios for node 9 are considered negative when the corresponding average values are calculated, as highlighted by the deep blue color.}}
\vspace{-2mm}
\label{Table:6}
\begin{center}
\begin{threeparttable}[b]
\setlength{\tabcolsep}{2.0mm}{
\begin{tabular}{c|ccccccccccccc}
\hline
 & Dis$.^1$(mm) & \multicolumn{2}{c}{(2,2)} & \multicolumn{2}{c}{(4,4)} & \multicolumn{2}{c}{(6,6)} & \multicolumn{2}{c}{(8,8)} & \multicolumn{2}{c}{(10,10)} &  &  \\ \hline
 & Nodes & \multicolumn{10}{c}{Error ratio(\%)} & A-Ave$.^2$(\%) & SD$.^4$(\%) \\
 & 2+3 & -1 & -6 & -4 & -3 & -4 & -5 & -6 & -6 & -6 & -7 & \cellcolor[HTML]{F9D7AE}4.80 & \cellcolor[HTML]{F9D7AE}1.72 \\
 & 4+5 & -8 & -6 & -6 & -4 & -4 & -6 & -5 & -6 & -6 & -6 & \cellcolor[HTML]{F9D7AE}5.70 & \cellcolor[HTML]{F9D7AE}1.10 \\
 & 6+7 & -3 & -4 & -3 & -5 & -5 & -3 & -5 & -4 & -6 & -5 & \cellcolor[HTML]{F9D7AE}4.30 & \cellcolor[HTML]{F9D7AE}1.00 \\
 & \cellcolor[HTML]{9B9B9B}8+9 & \cellcolor[HTML]{9B9B9B}-3 & \cellcolor[HTML]{9B9B9B}3 & \cellcolor[HTML]{9B9B9B}-1 & \cellcolor[HTML]{9B9B9B}6 & \cellcolor[HTML]{9B9B9B}-1 & \cellcolor[HTML]{9B9B9B}5 & \cellcolor[HTML]{9B9B9B}0 & \cellcolor[HTML]{9B9B9B}7 & \cellcolor[HTML]{9B9B9B}N.A$.^3$ & \cellcolor[HTML]{9B9B9B}N.A$.^3$ & \cellcolor[HTML]{9B9B9B}3.25 & \cellcolor[HTML]{9B9B9B}2.38 \\
\multirow{-6}{*}{Two-adjacent nodes} & A-Ave$.^2$(\%) & \cellcolor[HTML]{ECF4FF}3.75 & \cellcolor[HTML]{72B8EE}4.75 & \cellcolor[HTML]{ECF4FF}3.50 & \cellcolor[HTML]{72B8EE}4.50 & \cellcolor[HTML]{ECF4FF}3.50 & \cellcolor[HTML]{72B8EE}4.75 & \cellcolor[HTML]{ECF4FF}4.00 & \cellcolor[HTML]{72B8EE}5.75 & \cellcolor[HTML]{ECF4FF}6.00 & \cellcolor[HTML]{ECF4FF}6.00 & \cellcolor[HTML]{D9F5B1}4.51 & \cellcolor[HTML]{D9F5B1}\textbackslash{} \\ \hline
 & Dis$.^1$(mm) & \multicolumn{2}{c}{(2,10)} & \multicolumn{2}{c}{(4, 8)} & \multicolumn{2}{c}{(6, 6)} & \multicolumn{2}{c}{(8, 4)} & \multicolumn{2}{c}{(10, 2)} &  &  \\ \hline
 & Nodes & \multicolumn{10}{c}{Error ratio(\%)} & A-Ave$.^2$(\%) & SD$.^4$(\%) \\
 & 2+8 & -6 & -1 & -5 & -2 & -6 & -2 & -5 & -4 & -5 & -6 & \cellcolor[HTML]{F9D7AE}4.20 & \cellcolor[HTML]{F9D7AE}1.78 \\
 & 4+7 & -5 & -4 & -6 & -4 & -5 & -4 & -5 & -5 & -5 & -6 & \cellcolor[HTML]{F9D7AE}4.9 & \cellcolor[HTML]{F9D7AE}0.7 \\
\multirow{-4}{*}{Two-non-adjacent nodes} & A-Ave$.^2$(\%) & \cellcolor[HTML]{ECF4FF}5.50 & \cellcolor[HTML]{ECF4FF}2.50 & \cellcolor[HTML]{ECF4FF}5.50 & \cellcolor[HTML]{ECF4FF}3.00 & \cellcolor[HTML]{ECF4FF}5.50 & \cellcolor[HTML]{ECF4FF}3.00 & \cellcolor[HTML]{ECF4FF}5.00 & \cellcolor[HTML]{ECF4FF}4.50 & \cellcolor[HTML]{ECF4FF}5.00 & \cellcolor[HTML]{ECF4FF}6.00 & \cellcolor[HTML]{D9F5B1}4.55 & \cellcolor[HTML]{D9F5B1}\textbackslash{} \\ \hline
\end{tabular}}
\begin{tablenotes}
     \item \footnotesize{Dis$.^1$ denotes the displacement; A$.^2$ indicates the average value; N.A.$^3$ indicates that the FE method in Ansys fails; SD$.^4$ indicates the standard deviations.}
   \end{tablenotes}
\end{threeparttable}
\end{center}
\vspace{-6mm}
\end{table*}

\vspace{-5mm}
\subsection{Contact Force Estimation}
\vspace{-1mm}
The proposed approach can estimate the applied contact forces by the developed displacement-force modeling, \textbf{Algorithm 4}. In addition, the parameters of the soft gripper are shown in Table I, except for the node radius modification factor of 0.7 based on the above analysis. 
\subsubsection{Evaluation of the single-node cases}

\ 

% \begin{figure*}[thbp]
% \centerline{\includegraphics[width=1.8\columnwidth]{Fig/The deformations of a gripper for two-adjacent-node displacement loads.png}}
% \caption{\small{The deformations of a gripper for two-adjacent-node displacement loads.}}
% \label{Fig:12}
% \vspace{-3mm}
% \end{figure*}
% \begin{figure*}[thbp]
% \centerline{\includegraphics[width=1.9\columnwidth]{Fig/The deformations of a gripper for two-non-adjacent-node displacement loads.png}}
% \caption{\small{The deformations of a gripper for two-non-adjacent-node displacement loads at nodes 3, 9 (A) and 5,8. }}
% \label{Fig:13}
% \vspace{-5mm}
% \end{figure*}

\begin{figure}[htbp]
\vspace{-3mm}
\centerline{\includegraphics[width=0.95\columnwidth]{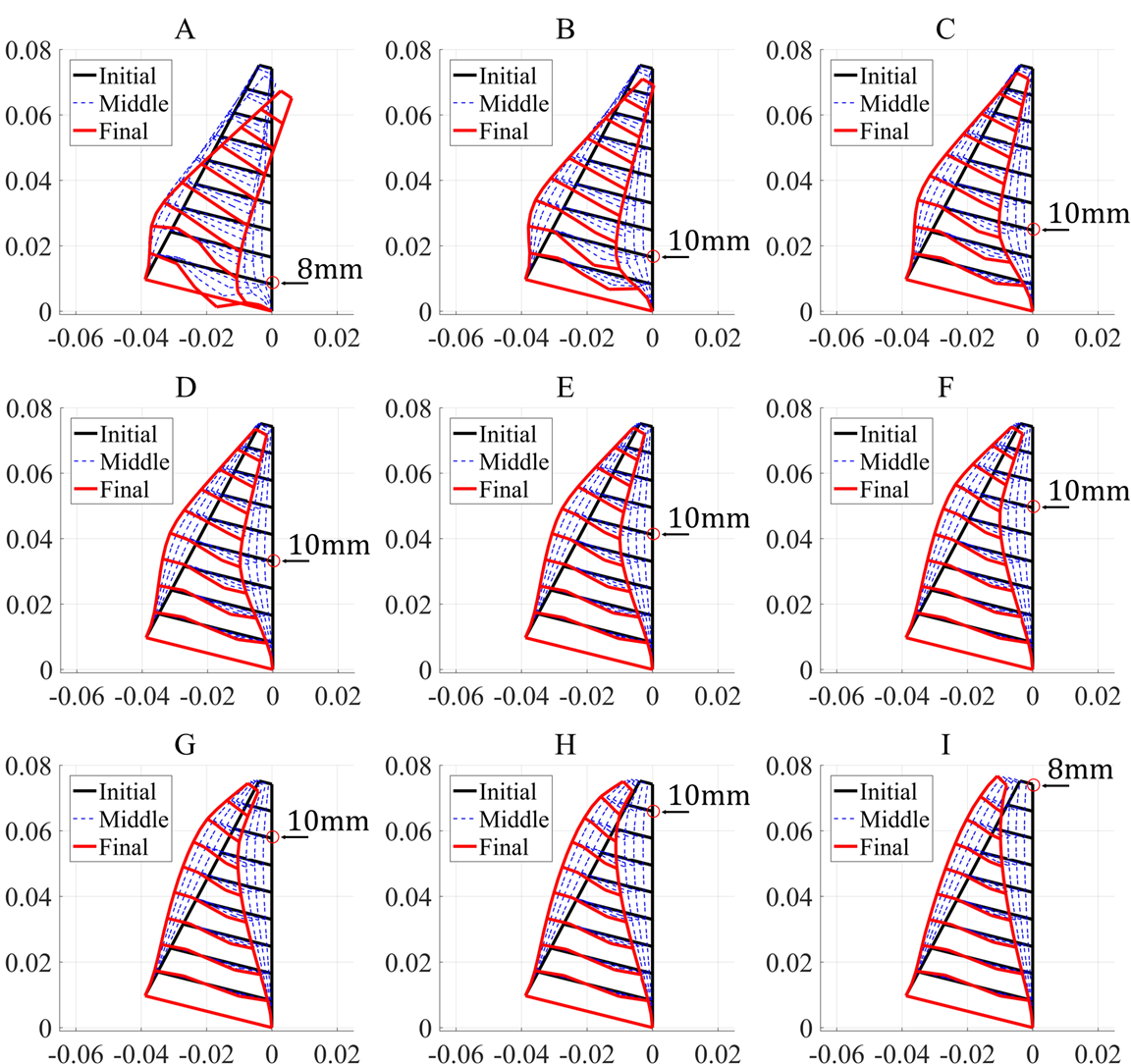}}
\caption{\small{The deformations of a gripper for single-node displacement loads. Displacements at nodes 1(A), 2(B), 3(C), 4(D), 5(E), 6(F), 7(G), 8(H), 9(I).}}
\label{Fig:11}
\vspace{-6mm}
\end{figure}

Here we compare the proposed method's ability to estimate force at single node. Each node is subjected to a horizontal external displacement load, as indicated in Fig. \ref{Fig:11}. A 10mm-displacement can be considered a large deformation with respect to the dimension of the gripper. For nodes 1 and 9, since the FEA simulation fail in estimating forces given the 10mm displacements, the maximum applied displacement is set to be 8mm (see Table II). The reaction force obtained from the proposed method is applied to an FEA simulation with 14278 elements and large deflection, and the resulting displacement is compared with the original displacement load from the mathematical model. For a quantitative comparison, Table II indicates the rates of displacement discrepancies/errors from compared results predicted by the proposed model and the FEA method, respectively. The overall average estimation error rate is 3.66\%. For the dense mesh model, displacement discrepancies are small \textcolor{DarkOrange}{(around 3\%-4\%, SD around 1\%)} for displacements between 2mm to 10mm with 2mm intervals, indicating that the proposed method can accurately analyze the displacement-to-force (estimate the contact forces from displacements) of a fin-ray finger. Surprisingly, the displacement discrepancy does not appear to be roughly proportional to the displacement at a node, suggesting the mathematical model can predict the large deformation without sacrificing accuracy. 

\begin{figure*}[thbp]
\centerline{\includegraphics[width=1.7\columnwidth]{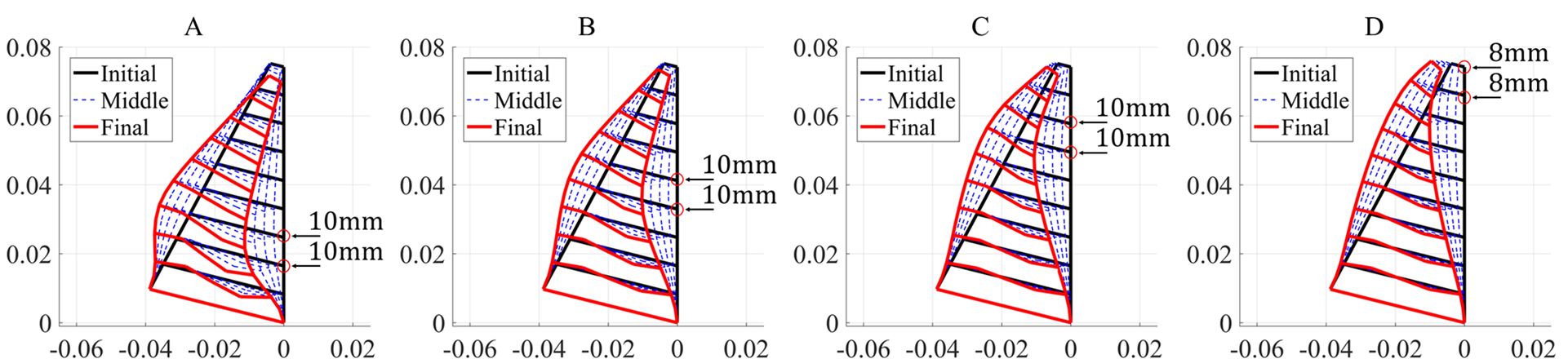}}
\caption{\small{The deformations of a gripper for two-adjacent-node displacement loads.}}
\label{Fig:12}
\vspace{-3mm}
\end{figure*}
\begin{figure*}[thbp]
\centerline{\includegraphics[width=1.65\columnwidth]{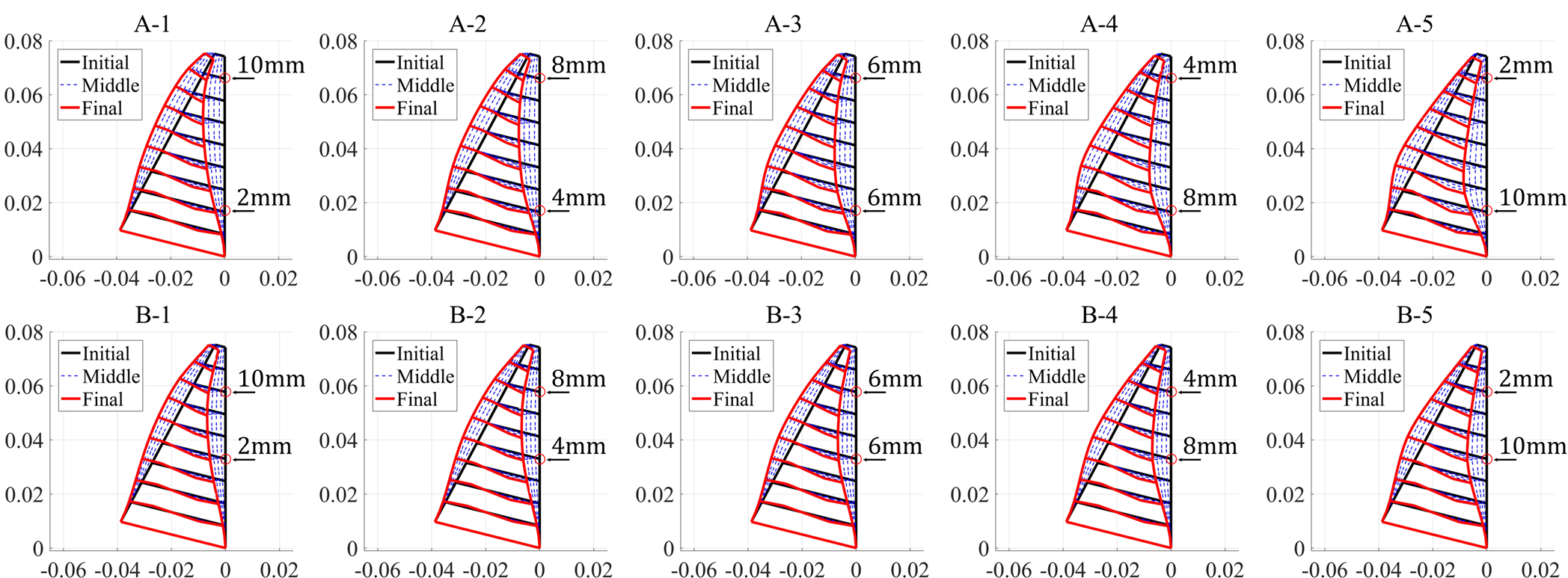}}
\caption{\small{The deformations of a gripper for two-non-adjacent-node displacement loads at nodes 3, 9 (A), and 5, 8 (B). }}
\label{Fig:13}
\vspace{-5mm}
\end{figure*}

\textcolor{Blue}{The FEA simulation in Ansys fail in estimating forces at the 10mm displacements at node 1 and node 9 (see Table II).} By contrast, the proposed model well handles these cases. Actually, error ratios at these nodes are not considered when calculating average values, as they are not usually used for contacting objects in gripper with fin-ray structure.  They differ slightly among most displacement-force cases. In particular, these error ratios are very close, within 6\% (Absolute value). Note that the proposed model demonstrates excellent consistency in terms of different positions and displacements. 
In particular, from node 3 to node 7, there are small fluctuations in displacement discrepancies at the same displacement. Moreover, displacements are gradually increased from 2mm to 10mm for all the nodes at the front beam; however, almost no obvious effect is generated on changing the displacement discrepancies/errors at the same node. At node 8, the proposed model illustrates an excellent performance with almost zero errors. 

% \begin{figure}[b]
% \vspace{-8mm}
% \centerline{\includegraphics[width=0.95\columnwidth]{Fig/The deformation of a gripper for three-non-adjacent-node.png}}
% \caption{\small{The deformations of a gripper for three-adjacent-node displacement loads.}}
% \label{Fig:14}
% \vspace{-2mm}
% \end{figure}

\subsubsection{Evaluation of more than one-node cases}

\ 

(1) Displacements at two nodes

To evaluate the performance of the proposed method in estimating contact forces at two nodes, we conduct experiments focusing on the two-adjacent-node and two-non-adjacent-node cases. The setup of the simulations is similar to the above subsection, except that those two horizontal displacements are applied simultaneously at two selected nodes, as shown in Fig. \ref{Fig:12} and Fig. \ref{Fig:13}. Statistical indicators of the force estimations from both the proposed model and the FEA method are illustrated in Table III. Therefore, it is reasonable to conclude that the proposed model can utilize finger deformations to accurately predict the contact forces when applied at two nodes. It is found that the proposed model can handle non-convergence at node 9 with 10mm displacement, while the FEA method fails, indicating the advantage of the proposed model over the FEA method. 

\begin{figure}[t]
\vspace{-0mm}
\centerline{\includegraphics[width=\columnwidth]{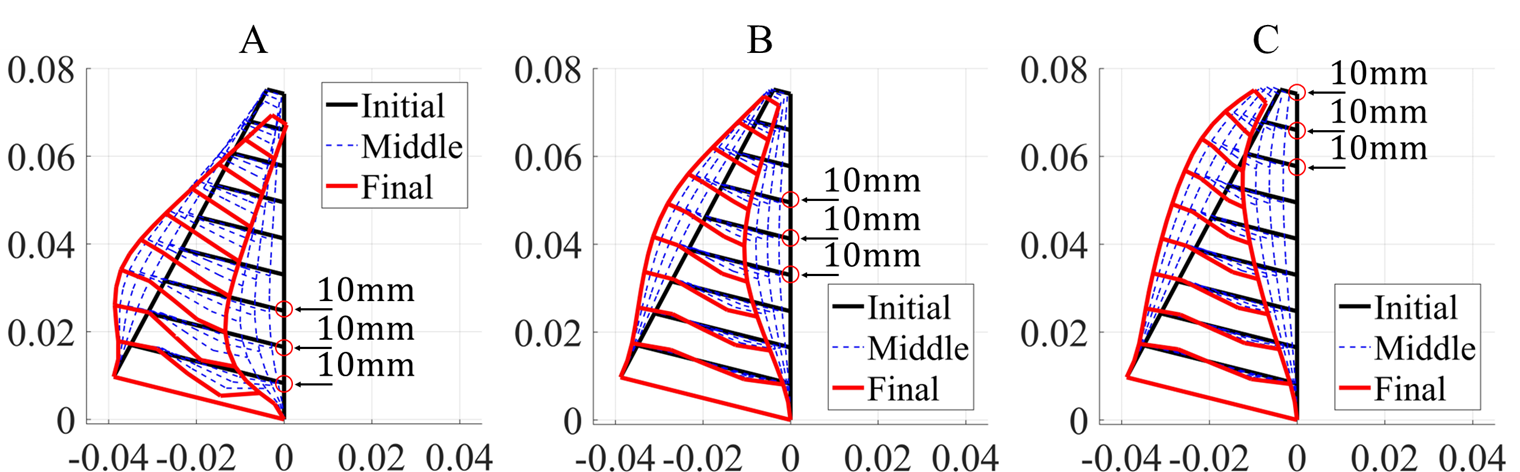}}
\caption{\small{The deformations of a gripper for three-adjacent-node displacement loads.}}
\label{Fig:14}
\vspace{-8mm}
\end{figure}

Regardless of two-adjacent-node and two-non-adjacent-node cases, the proposed method demonstrates a similar performance to the single-node case on the estimation accuracy, mostly maintaining around 4\%-6\% (absolute value) estimation errors for different positions and displacements, \textcolor{DarkOrange}{and SD is around 1\%.} The reason is that the proposed model considers stiffness rather than a specific position. In terms of the two-adjacent-node cases, the corresponding average error ratios from the first group to the third group (i.e., from base to tip) are around 5\% (see Fig. \ref{Fig:12} and Table III). 
% Note that it is unavailable for the FEA method to predict the contact force applied at node 8 and node 9 for the 10mm displacement.
Moreover, in terms of two-non-adjacent-node cases, the discrepancy/error ratios at nodes (2 and 8) and (4 and 7) are less than 6\% (see Fig. \ref{Fig:13} and Table III). To sum up, the proposed approach is capable of sensing contact forces exerted on its middle part but has a poor performance when the contact points are near the tip or base, especially the tip. Moreover, the nonlinearity of the compliant finger results in a systemic error. For instance, the calibrated Young’s modulus has a difference from the real one, which is a potential reason why the proposed model has a poor performance in estimating contact forces at the finger base and tip. 

(2) Displacements at three nodes

\begin{figure*}[th]
\vspace{-1mm}
\centerline{\includegraphics[width=1.8\columnwidth]{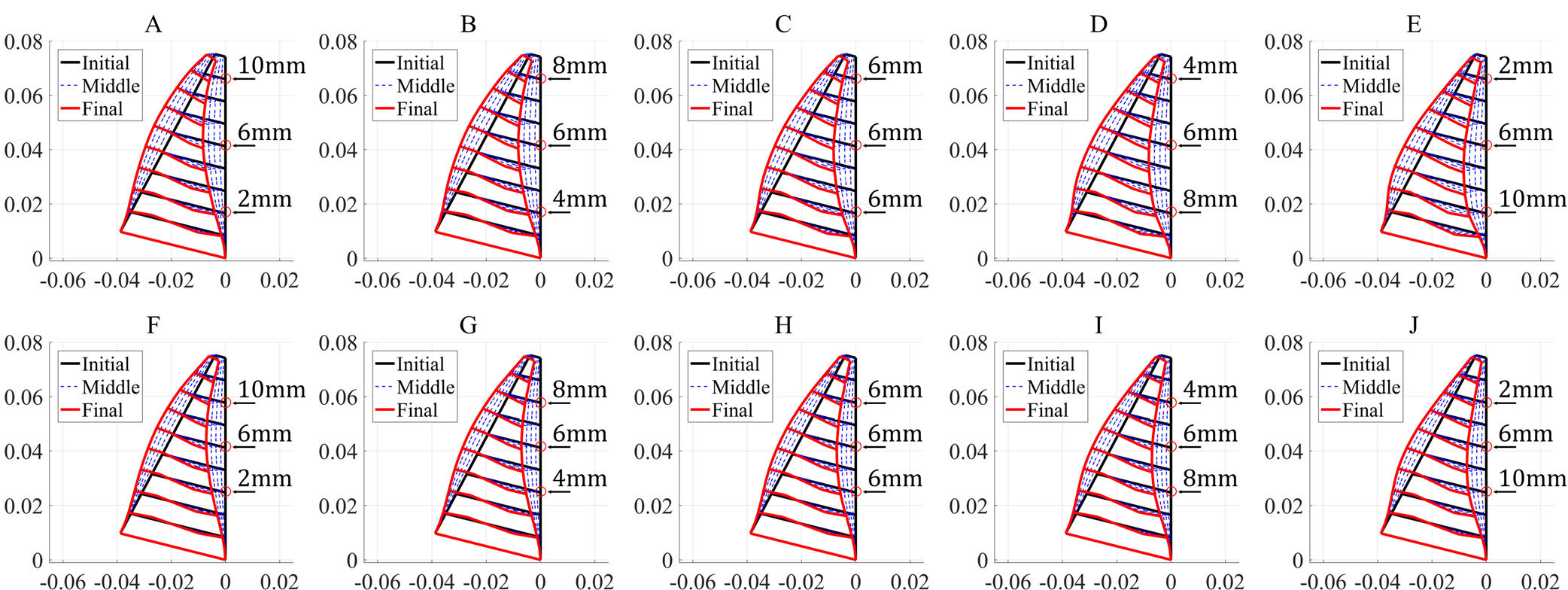}}
\vspace{-1mm}
\caption{\small{The deformations of a gripper for three-non-adjacent-node displacement loads.} }
\label{Fig:15}
\vspace{-3mm}
\end{figure*}

We test the proposed approach in estimating forces at multiple nodes, focusing on three-node cases, including adjacent and non-adjacent nodes. The simulation comparison is similar, where three pre-defined horizontal displacement loads are applied at selected nodes simultaneously, and their reaction forces are recorded, as illustrated in Fig. \ref{Fig:14} and Fig. \ref{Fig:15}.
% The force values are then passed to the standard FEA simulation, and the resulting deformation values are recorded, and their discrepancies with analytical results are compared.
The comparison result suggests the average sensing error ratios are around 5\%, \textcolor{DarkOrange}{the SD within 3\%,} which illustrates that the proposed model can well estimate forces at multiple nodes (see Table IV). Moreover, for force estimation on different positions and displacements, the proposed model shows excellent consistency, which is similar to the single-node cases.

The proposed model presents an unstable phenomenon in the first three-adjacent-node case, with the displacements gradually increasing. The high sensing error ratios at node 1 of the 8mm and 10mm displacements exceed 10\% (absolute value). While for the middle parts (node 3 to node 9), the average error ratios are within 5\%. One possible reason is that the joint points of the crossbeam and the beam have almost unchanged deformations at the finger base. \textcolor{blue}{To complete a precise pinch, high accuracy at the fingertip is necessary. However, due to the compliant nature of these grippers, they are more employed to compliantly envelop  the target objects, with the middle nodes first making contact and the peripheral nodes adaptively approaching to complete the grasp.} In practical applications, inaccurate estimations at the base of the finger, where contact with the object is infrequent, will not significantly affect the actual grasp prediction.
% Thus, inaccurate estimations at the compliant finger base do not generate a vital effect on practical grasp applications.
The estimating capability becomes better for estimating forces at three-non-adjacent-node cases without exceeding 5\%, as shown in Table IV. 
\textcolor{Yellow}{The overall average error for both adjacent and non-adjacent nodes is within 5\%, compared to 8\% reported by Xu\cite{ref6}}. 
The study demonstrates that the proposed model maintains good accuracy in estimating the contact forces with respect to large objects.

\begin{table*}[th]
\renewcommand\arraystretch{0.8}
\vspace{-1mm}
\caption*{\small{Table \uppercase\expandafter{\romannumeral4}. Results of the evaluations of three-node force estimation. The color notations are the same as Table III.}}
\vspace{-4mm}
\label{Table:7}
\begin{center}
\begin{threeparttable}[b]
\setlength{\tabcolsep}{2.1mm}{\begin{tabular}{c|cccccccccccccccccc}
\hline
 & Dis$.^1$(mm) & \multicolumn{3}{c}{(2,2,2)} & \multicolumn{3}{c}{(4,4,4)} & \multicolumn{3}{c}{(6,6,6)} & \multicolumn{3}{c}{(8,8,8)} & \multicolumn{3}{c}{(10,10,10)} &  & \multicolumn{1}{l}{} \\ \hline
 & Nodes & \multicolumn{15}{c}{Error ratio(\%)} & A-Ave$.^2$(\%) & \multicolumn{1}{l}{SD$.^5$(\%)} \\
 & 1+2+3 & -8 & -3 & -5 & -6 & -6 & -6 & -9 & -7 & -7 & -12 & -8 & -9 & -15 & -10 & -10 & \cellcolor[HTML]{F9D7AE}8 & \cellcolor[HTML]{F9D7AE}3 \\
 & 4+5+6 & -7 & -6 & -4 & -6 & -4 & -4 & -4 & -5 & -6 & -5 & -5 & -6 & -5 & -6 & -6 & \cellcolor[HTML]{F9D7AE}5 & \cellcolor[HTML]{F9D7AE}1 \\
 & 7+8+9 & -3 & 0 & 4 & -3 & 0 & 4 & -3 & 0 & 4 & -3 & -1 & 4 & -3 & -1 & 4 & \cellcolor[HTML]{F9D7AE}2 & \cellcolor[HTML]{F9D7AE}3 \\
\multirow{-5}{*}{3AN$^3$} & A-Ave$.^2$(\%) & \cellcolor[HTML]{ECF4FF}6 & \cellcolor[HTML]{ECF4FF}3 & \cellcolor[HTML]{72B8EE}4 & \cellcolor[HTML]{ECF4FF}5 & \cellcolor[HTML]{ECF4FF}3 & \cellcolor[HTML]{72B8EE}5 & \cellcolor[HTML]{ECF4FF}5 & \cellcolor[HTML]{ECF4FF}4 & \cellcolor[HTML]{72B8EE}6 & \cellcolor[HTML]{ECF4FF}7 & \cellcolor[HTML]{ECF4FF}5 & \cellcolor[HTML]{72B8EE}6 & \cellcolor[HTML]{ECF4FF}8 & \cellcolor[HTML]{ECF4FF}6 & \cellcolor[HTML]{72B8EE}7 & \cellcolor[HTML]{D9F5B1}5 & \cellcolor[HTML]{D9F5B1}\textbackslash{} \\ \hline
 & Dis$.^1$(mm) & \multicolumn{3}{c}{(2,6,10)} & \multicolumn{3}{c}{(4,6,8)} & \multicolumn{3}{c}{(6,6,6)} & \multicolumn{3}{c}{(8,6,4)} & \multicolumn{3}{c}{(10, 2)} &  & \multicolumn{1}{l}{} \\ \hline
 & Nodes & \multicolumn{15}{c}{Error ratio(\%)} & A-Ave$.^2$(\%) & \multicolumn{1}{l}{SD$.^5$(\%)} \\
 & 2+5+8 & -6 & -6 & -2 & -6 & -6 & -3 & -5 & -6 & -4 & -5 & -6 & -5 & -5 & -7 & -7 & \cellcolor[HTML]{F9D7AE}5 & \cellcolor[HTML]{F9D7AE}1 \\
 & 3+5+7 & -5 & -5 & -4 & -6 & -5 & -5 & -5 & -5 & -5 & -5 & -5 & -5 & -5 & -5 & -6 & \cellcolor[HTML]{F9D7AE}5 & \cellcolor[HTML]{F9D7AE}0 \\
\multirow{-4}{*}{3NAN} & A-Ave$.^2$(\%) & \cellcolor[HTML]{ECF4FF}6 & \cellcolor[HTML]{ECF4FF}6 & \cellcolor[HTML]{ECF4FF}3 & \cellcolor[HTML]{ECF4FF}6 & \cellcolor[HTML]{ECF4FF}6 & \cellcolor[HTML]{ECF4FF}4 & \cellcolor[HTML]{ECF4FF}5 & \cellcolor[HTML]{ECF4FF}6 & \cellcolor[HTML]{ECF4FF}5 & \cellcolor[HTML]{ECF4FF}5 & \cellcolor[HTML]{ECF4FF}6 & \cellcolor[HTML]{ECF4FF}5 & \cellcolor[HTML]{ECF4FF}5 & \cellcolor[HTML]{ECF4FF}6 & \cellcolor[HTML]{ECF4FF}7 & \cellcolor[HTML]{D9F5B1}5 & \cellcolor[HTML]{D9F5B1}\textbackslash{} \\ \hline
\end{tabular}}
\begin{tablenotes}
     \item \footnotesize{Dis$.^1$ denotes the displacement; A-Ave$.^2$ indicates around average values; 3AN$^3$ represents three-adjacent nodes; 3NAN$^4$ represents three-non-adjacent nodes; SD$.^5$ indicates the standard deviations.}
   \end{tablenotes}
\end{threeparttable}
\end{center}
\vspace{-7mm}
\end{table*}
\vspace{-2.5mm}
\section{Discussion}
\subsection{Advantages and limitations of the proposed Displacement-Force modeling}
Our proposed methods is not a substitute or supplement for existing methods. \textcolor{Purple}{The existing commercial FE software is designed to meet structural design optimization. Though it is good at detailed simulation, too much variables introduced would cause confusions for researchers and the key design index parameters are not intuitive. In addition, they have low iteration speed and is not suitable for application in the closed-loop control for the robot. By contrast, the proposed model is characterised with fair degree of accuracy and efficiency in displacement-force estimation. } 

\textcolor{Green}{Besides, the proposed method provides a general mathematical model depicting displacement-force relationship and can be further extended to more soft/rigid grippers. Especially, a soft/continuum gripper can be considered a Fin-Ray gripper with numerous crossbeams and low stiffness material for readily deforming. While a rigid gripper may be regarded as a Fin-Ray gripper with high-stiff material. Besides, unlike existing works, where a lot of assumptions are made to express axial/rotational deformation and the connection types, such as in Shan's model, they assume the crossbeams as inextensible beams and ignore their axial deformation\cite{ref4}. Thanks to the flexible nature of co-rotational methods, we could well handle the above constraints with fewer assumptions.}

\textcolor{Red}{Besides, the proposed model has the following limitations. The sources of differences between the different nodes can be ascribed to the following threefold aspects. First, the discretization degree based on the number of nodes meshing the model causes the estimation error of the stiffness. Second, the shapes of the crossbeams tend to be ‘S’-shaped when undergoing large deformations of the finger, which brings in inaccurate estimations for stiffness matrixes in the model. Third, the deformation at the base (node 1) of the beam structure of compliant lengths generates a significant deformation compared with other locations, but it is difficult to obtain a precision measurement.}

\subsection{Influence of the results on Physical Experiments}
\textcolor{Blue}{Although our experiments are simulated by FEA, the proposed model is able to be applied in a physical model. In FEA, we mentioned some physical parameters, such as the modification factor $\mu$, and Young's modulus, which not only have a large impact on the simulation results, but are also crucial in physics experiments. If physical experiments are performed, these physical parameters can be easily obtained from the physical model by measurement.}
\

\vspace{-2mm}
\section{conclusions}

In Part II of this article, detailed derivations of the displacement-force model have been devised to illustrate the intrinsic force-sensing principles behind the nonlinear deformation of grippers. We further explore the influence of modeling parameters on the algorithm's performance in accuracy, and results indicate that a denser mesh and a bigger node radius factor are preferred. Meanwhile, the comparison with FEA simulations validates that the proposed displacement-force model can accurately and efficiently estimate the loaded forces, with an average error $<5\%$, regardless of various design parameters. 

% Part I of this paper introduces the fundamental theory of the co-rotational approach, together with a force-displacement mapping providing gripper deformation under external forces.

To sum up, this paper innovatively devises, demonstrates, and experimentally verifies a universal theoretical model that mutually depicts the bidirectional relationship between the gripper's displacement and contact forces, facilitating both the optimization of gripper design and compliant grasping in a force-aware manner, sensor-free. Especially from the perspective of theoretical modeling, this work lays a solid foundation and provides a theoretical background for the development of more compliant grippers, providing a promising mathematical tool for the control of soft fin-ray grippers and represents a first step toward a more effective design of more compliant grippers.  

Future work will focus on the grasping experiments of the fin-ray gripper, where a gripper with two fin-ray-like fingers will be constructed, equipped with a camera that can capture shape deformations, to physically evaluate the performance and further validate the accuracy of the proposed model.
% This paper innovatively devises, demonstrates and experimentally verifies a universal theoretical model that mutually depicts the bidirectional relationship between gripper's displacement and contact forces, facilitating both the optimization of gripper design and compliant grasping in a force-aware manner, sensor-free. Part I of this paper introduces the fundamental theory of the co-rotational approach, together with a force-displacement mapping providing gripper deformation under external forces. Particularly in this part, detailed derivations of displacement-force model are implemented to illustrate the intrinsic force sensing principles behind the nonlinear deformation of grippers. We further explore the influence of modeling parameters on algorithm's performance in accuracy. Results of the simulation experiments have shown that the proposed displacement-force model can accurately and efficiently estimate the loaded forces, with an average error $<5\%$ compared with FEA simulations, regardless of various design parameters. From the perspectives of theoretical modeling, this work  lays a solid foundation and provides a theoretic background for the development of more compliant grippers, providing a promising mathematical tool for control of soft fin-ray grippers and represents a first step towards more effective design of more compliant grippers. 

\addtolength{\textheight}{+0.5cm}   % This command serves to balance the column lengths
\AtNextBibliography{\small}
\printbibliography
% \vfill

\end{document}